\documentclass[letterpaper, 10 pt, conference]{ieeeconf}  

\IEEEoverridecommandlockouts                              

\everypar{\looseness=-1}
\usepackage{times}  
\usepackage{helvet} 
\usepackage{courier}  
\usepackage[hyphens]{url}  
\usepackage{graphicx} 
\usepackage{caption} 
\usepackage{latexsym}

\usepackage{microtype}

\usepackage{multicol}

\newcommand{\E}[1]{\ensuremath{\mathbb{E}\left[#1\right]}}

\usepackage{amssymb}
\usepackage{amsmath}

\usepackage{graphicx}

\usepackage{cleveref}
\crefname{section}{s}{ss}
\crefname{section}{s}{ss}
\crefname{table}{Table}{}
\crefname{figure}{Figure}{}
\crefname{algorithm}{Alg.}{}
\crefname{equation}{Eq.}{}
\crefname{appendix}{Appendix}{}
\crefformat{section}{\S#2#1#3}  

\usepackage{multirow}
\usepackage{color}
\usepackage{soul}			
\usepackage[hang, flushmargin]{footmisc} 
\usepackage{array}			
\usepackage{balance}		
\usepackage{xspace}			
\usepackage{lipsum}
\usepackage[flushmargin]{footmisc}
\usepackage{quotes}
\let\labelindent\relax
\usepackage[shortlabels,inline]{enumitem}

\newcolumntype{L}[1]{>{\raggedright\arraybackslash}m{#1}}
\newcolumntype{C}[1]{>{\centering\arraybackslash}m{#1}}
\usepackage{enumitem}
\usepackage{wrapfig}

\usepackage[switch]{lineno} 

\newcommand{\Concept}[1]{\textit{#1}}

\newcommand{\pseudosection}[1]{\vspace{1ex}\noindent \textbf{{#1}}}

\everypar{\looseness=-1} 
\title{\LARGE \bf
Neural Variational Learning for Grounded Language Acquisition
}

\author{Nisha Pillai, Cynthia Matuszek, and Francis Ferraro
\thanks{*This material is based in part upon work supported by the National Science Foundation under Grant Nos. 1657469, 1940931, and 2024878.}
\thanks{$^{1}$All authors are with the University of Maryland, Baltimore County, 1000 Hilltop Circle, Baltimore, MD, 21250, USA. (npillai1, cmat, ferraro@umbc.edu)}}

\date{}

\begin{document}
\maketitle




\begin{abstract}
We propose a learning system in which language is grounded in visual percepts without specific pre-defined categories of terms. We present a unified generative method to acquire a shared semantic/visual embedding that enables the learning of language about a wide range of real-world objects. We evaluate the efficacy of this learning by predicting the semantics of objects and comparing the performance with neural and non-neural inputs. We show that this generative approach exhibits promising results in language grounding without pre-specifying visual categories under low resource settings. Our experiments demonstrate that this approach is generalizable to multilingual, highly varied datasets.
\end{abstract}

\section{Introduction}
\label{sec:intro}


In grounded language theory, the semantics of language are defined by how symbols connect to the underlying real world---the so-called ``symbol grounding problem''~\cite{harnad1990symbol}. Symbol grounding requires deep understanding at multiple levels, from the very high, such as a robot navigating by following instructions~\cite{chi2020aaai} or a system that can generate a text narrative from a photo album~\cite{wang2019hierarchical}, to the very low, such as a system that can understand individual traits and characteristics of everyday objects~\cite{sinapov2016IJCAI}. %
While joint vision/language problems have become increasingly popular within NLP, with developments in tasks like visual question answering~\cite[i.a.]{Antol_2015_ICCV}, the core ``symbol grounding problem'' remains unsolved.

Meanwhile, as robots become more capable and affordable, the idea of deploying them in human-centric environments becomes more realistic. Such robots, in turn, need to be able to take instructions from people in a natural, intuitive way, including those pertaining to---that is, grounded in---the specific environment in which they are operating. One particular problem is the ability to understand and predict \textit{concepts} for various items. As classically defined~\cite{MatuszekICML2012}, learning these concepts amounts to learning classifiers that can predict whether particular input modalities or sensor readings can be linguistically and notionally referred to by a particular label---a ``concept.'' These concepts are traditionally learned based on language, and often (although not always) refer to specific attributes and/or types of attributes of an item, such as object type, material~\cite{richards2020IROS}, weight, or sound~\cite{bisk2020experience}.

We focus on this concept learning problem, and specifically on removing the assumption that only concepts in pre-defined categories are to be learned. Object percepts are collected using an RGB-D sensor, and natural language descriptions are provided by crowdsource workers. Using this data, we examine a \textit{computationally reasonable} visual pre-training approach that improves on existing concept learning systems; is robust to modality featurization/embedding; and performs well in low-data settings. The variational autoencoder approach we explore in this paper has the benefits of simplicity and approachability, while still demonstrating effectiveness in the low-resource setting we consider.

\begin{figure*}[bt]
    \centering	
   \includegraphics[width=0.77\textwidth]{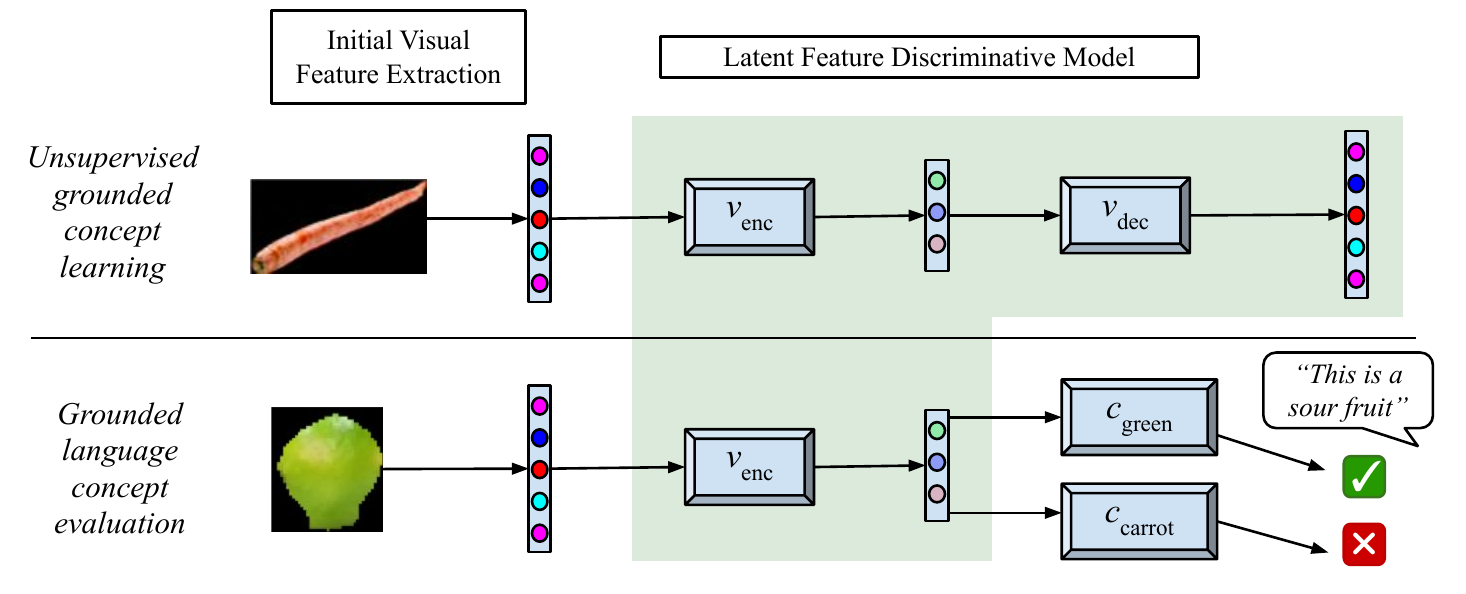}    
	\caption{Design diagram of concept grounding using an unsupervised latent feature discriminative method. For every object, we extract visual features and learn a representative feature embedding by applying a latent feature discriminative model. The visual variation encoder ($v_{enc}$) embeds the cumulative visual features to low-dimensional feature representation, and the visual variational decoder ($v_{dec}$) decodes the embedding to the visual features. The extracted low-dimensional feature embeddings are then used to create a concept classifier ($c_{concept}$) for language grounding.}
    \label{fig:diagram1}
\end{figure*}

We generalize language acquisition by using novel, generally applicable visual percepts with natural descriptions of real-world objects. Instead of creating classifiers for a fixed set of high-level object attributes, we use a combination of features to create a general classifier for terms in language. Deep generative models are used to obtain a representative unified visual embedding from the combination of visual features (see~\cref{fig:diagram1}).

Our core contribution is \textbf{a mechanism for generalizing language acquisition with an unsupervised neural variational autoencoder, which relies only on small amounts of data and requires no pre-trained image models.}  In order to compare to existing work, we evaluate against learning concepts in predefined categories; however, the work presented here does not rely on such categories. Our VAE provides comparable results to previous work that uses a fixed set of concept types, demonstrating that it is possible to learn more generalized language groundings. We also demonstrate consistent improvements in Spanish and Hindi grounded language understanding.

\section{Related Work}\label{sec:related}

Our focus is on the symbol grounding problem, rather than symbol emergence~\cite{doi:10.1080/01691864.2016.1164622},
which aims to account for the dynamics of symbol systems in society. Our experiments are designed to learn attributes~\cite{BergBergShih2010, Kery2019ROMAN, PillaiRSS2016workshopActive} from the noisy descriptions of real-world objects without specifying categories, whereas~\cite{paul2016efficient} and \cite{tellex2011approaching} propose to ground spatial concepts,~\cite{brawer2018situated} learns speech joining with context, and~\cite{Akula2020WordsAE} grounds natural language referring expressions for objects in images. 
Learning visual attributes such as color and shape is critical in robot object grasping~\cite{rao2018learning, levine2018learning} and manipulation tasks. 
Some studies ground language by partitioning feature space by context~\cite{thomason2016IJCAI}, whereas we intend to learn concepts without manually specifying attribute types. Our principal objective is to learn the concepts/characteristics of real-world objects from human annotations; as such, retrieving unknown objects~\cite{Nguyen2020RobotOR} from natural queries, learning semantic relations between words, and predicting missing categories are beyond the scope of this work. 

Deep learning has been successfully applied to many applications~\cite{Chen_2020_CVPR, pitti2021brain}. However, despite ongoing developments in zero-shot~\cite{Liu2021TaskAG} and few-shot learning~\cite{boudiaf2020information}, idiosyncratic scenarios---such as those encountered in households---generally still require large datasets, as do the pre-trained language representations~\cite{DBLP:conf/naacl/DevlinCLT19, lu2019vilbert} used in visio-linguistic learning. Our objective is to gain better prediction with a smaller, natural, and likely noisy dataset. A similar visio-linguistic application,  image captioning~\cite{lei2020mart} produces a sentence or captions of the content of an image. In contrast, our research focuses on learning about objects that a robot finds in its world. While image captioning aims to learn a scene in breadth, our work attempts to understand an object in-depth and understand traits of that object. 

Our architecture predicts visual percepts associated with language by training the representative latent probability distribution generated from cumulative visual features using a deep generative model. We employ autoencoders, which have been successful in a tasks such as 
3D shape analysis~\cite{Tan_2018_CVPR}, 
linguistic descriptions of robot actions~\cite{yamada2018paired}, and scene-text recognition~\cite{Qiao_2020_CVPR}, to generalize language acquisition. Our architecture does not separate attribute types while learning, but \cite{silberer2014learning} train stacked autoencoders for every modality by treating them separately and fusing them at the last layer to obtain meaningful representation. 

While LSTM based frameworks~\cite{rohrbach2016grounding} effectively ground textual phrases in images, we intend to show how a simple, efficient autoencoder can learn semantics from noisy, natural human annotation. Our research is most similar to that of \cite{PillaiAAAI2018}, in which language is learned by jointly connecting with visual characteristics of real world objects. We learn visual attributes~\cite{nyga2017no} from a small dataset, similar to few-shot learning~\cite{Hariharan_2017_ICCV, vinyals2016matching}, 
which learns from a few samples~\cite{liu-etal-2019-second, sun-etal-2019-hierarchical}, and zero-shot learning~\cite{lampert2014attribute, Han_2020_CVPR, Wang_2020_CVPR}.

\section{Approach}\label{sec:approach}
We suggest an effective generic visual classifier for training real-world object features with the noisy natural human descriptions. To learn the language and its association with visual perception, we extract a latent semantic embedding from the cumulative visual data and join it with the linguistic concepts. A high-level view of our approach can be formulated as follows: 1) Extract visual features that are associated with the perception; 2) Join all extracted visual features; 3) Use the latent feature discrimination method~\cite{kingma2014semi} 
based on an unsupervised neural variational autoencoder to extract meaningful, representative latent embedding from the cumulative feature set; and 4) Learn a general visual classifier (supervised) using the latent embedding created from the cumulative feature set (see~\cref{fig:diagram1}). Here we intend to show how this simple discriminative method is effective in generalizing visual classifiers. We describe our model and data corpus in section~\cref{subsec:embedding} and~\cref{subsec:data}.

\subsection{Unified Discriminative Learning Model}\label{subsec:embedding}
Our objective is to associate linguistic concepts, $W$, with the set of real-world objects, $O$, especially when we cannot assume there is a large amount of training data available. To learn this grounded association, we create a generalized visual feature embedding out of the features extracted from the object instances and use it to train a general classifier. Components of the unified discriminative model are below. 

\pseudosection{Straight-forward Pre-training for a VAE} We define $X$ as the feature vector extracted from the object $o$. In the experiments where we use attribute-based visual features, $X$ is $\langle f_1,f_2...f_n \rangle$ and $f_i$ is the type of visual feature extracted. Inspired by discriminative variational autoencoding~\cite{kingma2014semi}, we construct a representative, meaningful low-dimensional embedding, taking the cumulative feature vector $X$ as input.  We use a deep generative autoencoder that provides latent feature embedding for training grounded concept classifiers. While at this point variational autoencoders are a well-known tool, we argue that this use is a core strength of this paper's approach.

This variational autoencoder consists of an encoder, a decoder, and a loss function (see~\cref{fig:diagram1}). The encoder is a neural network that translates input data $X$ into latent variables $Z$. The encoder uses a neural network, $q_{\theta}(Z)$, to approximate $P(Z|X)$, our generic latent representation. Another network,  $P_{\phi}(X|Z)$---the decoder---is used to reconstruct $X$ from the latent variables $Z$. In this research, our challenge is to find an efficient representation of our feature space $P(Z|X)$ from the limited input data for our grounded learning task.

We approximate the posterior probability using a Gaussian function $q_{\theta}(z|x) = \mathcal{N}(z|\mu_{\theta}(x), \mathrm{diag}(\sigma^{2}_{\theta}(x)),$ %
where $\sigma^{2}(x)$ is a vector of standard deviations, $\mu(x)$ is a vector of means, both learned via multilayer perceptrons (MLPs). %
These are learned by minimizing the standard VAE loss:
\begin{equation} \label{eqn:lossfn}
L = - \E{\log p(x|z)} + \mathrm{KL}(q(z|x)  ||  p(z)).
\end{equation}
This loss is the sum of the reconstruction error (expectation of negative log-likelihood) and the KL divergence of approximation function and prior distribution ($KL(q(z|x) || p(z))$).

We define vectors of the mean $\mu(x)$ and standard deviation $\sigma^2(x)$---extracted from the variational autoencoder network---as the latent embedding $Z$. Employing an encoder function represented by a neural network, we learn the encoder weights of the unified discriminative model (UDM) by applying all the training data as input (see~\cref{fig:diagram1}).

\pseudosection{Category-free Visual Classifiers} For every relevant concept $w$, we learn a binary classifier $P(y_w = 1 | Z)$ for the positive items and $P(y_w = 0 | Z)$ for the negative items (see sample selection: section~\cref{samples}). $Z$ is defined as the visual groundings generated from the cumulative features. Unlike previous approaches, instead of creating a concept-per-attribute classifiers, we learn a single concept classifier. For example, instead of learning a ``red-as-color'' classifier by training on color features (alone), we create a unified general classifier for the concept ``red'' by associating a generalized probability distribution made from the visual features extracted from the perceived objects. We use binary logistic regression classifier to learn concept classifiers.

\subsection{Data Corpus}\label{subsec:data}

We demonstrate the utility of the UDM approach on two different, real robotics datasets. Both publicly available datasets contain vision and depth inputs (RGB-D) of objects collected during robot-world experiences, using sensors mounted on a robot. The first contains color and depth images of real-world objects in 72 categories~\cite{PillaiAAAI2018,lai2011dataset} which are divided into 18 classes. Objects include food objects such as `potato,' `tomato,' and `corn,' as well as children’s toys in several shapes such as `cube' and `triangle.' There are an average of 4.5 images collected for every object and 22 concepts to predict. 
The second dataset is an extension of the well-known UW RGB-D object set with 300 objects in 51 categories~\cite{lai2011dataset,richards2020IROS}, with 122 concepts.

\begin{figure}[tb]
\centering
\includegraphics[width=\columnwidth]{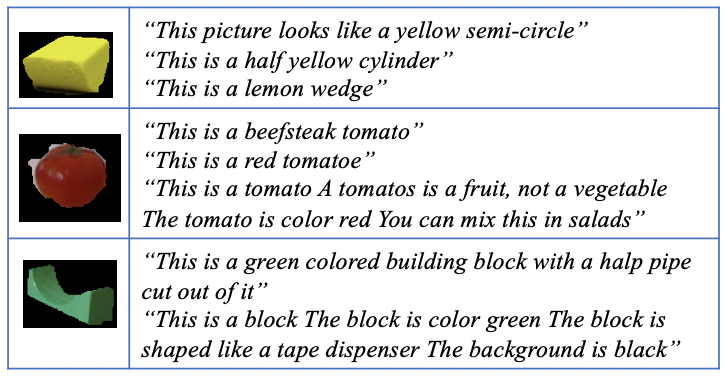}
	\caption{Object samples and language descriptions collected from Amazon Mechanical Turk annotators.  These descriptions are noisy, containing typographical errors.}
    \label{fig:dataset_amt}
\end{figure}

To learn linguistic concepts from natural, untrained human language, we gathered descriptions of these objects obtained via Amazon Mechanical Turk (see~\cref{fig:dataset_amt}), tokenized them, created one visual classifier per concept, and used them to learn real-world objects. Here, instead of building classifiers within specific categories, we create and learn a single visual classifier per term from a single general set of features (e.g., instead of learning separate possible classifiers such as both ``cube-as-shape'' and ``cube-as-object'' classifiers, we learn a single ``cube'' classifier). This work is similar to approaches in which language grounding is treated as an association of linguistically-based concepts with the visual percepts extracted from real-world objects~\cite{PillaiAAAI2018}. 

\section{Experimental Results}\label{subsec:experiments} 

In \cref{sec:expt:specification}, we detail the preprocessing steps for the training data and instantiation of the UDM model. In \cref{sec:expt:setup}, we describe the baselines, evaluation metric, and cross-fold setup.

\subsection{UDM Specification}
\label{sec:expt:specification}

\pseudosection{Initial Visual Features}\label{visual}
While the UDM VAE learns and computes refined embeddings, we experiment with and provide the VAE three different initial visual embeddings. %
In the first case, we use the same 703 visual features (averaged RGB values and kernel descriptors from associated depth images~\cite{BoLaiRenEtAl2011}) for each image that previous work has used~\cite{PillaiAAAI2018}. Kernel descriptors extract size, 3D shape, and depth edge from the RGB-D depth channel, and are efficient in shape and object classification. While this set of features does not use a neural network, we experiment with them because they are a proven and reliable set of features within the robotic-based vision and language processing~\cite{MatuszekICML2012}. %

We also examine neural image processing approaches (with pretrained ImageNet~\cite{deng09} weights) to demonstrate the generalizability and flexibility of the VAE. In the second case, we extraced a 1,024 dimension feature vector from SmallerVGGNet, a variant of the popular and strong object classifier VGGNet neural architecture \cite{DBLP:journals/corr/SimonyanZ14a}. However, as Neural Architecture Search Network (NASNetLarge)~\cite{zoph2018learning} obtained better top-1 and top-5 accuracy on multiple datasets compared to popular architectures like ResNet, Inception, and VGGNet, we examine it as the third case and extract 1,024 dimensional vectors as well. 


\pseudosection{Sample Selection}\label{samples}
We select positive object instances for every meaningful concept identified. We consider an object instance a `positive' example if the object is described by the concept's corresponding lexical form in any of that object's descriptions. If an instance makes use of a novel concept, we create a new visual classifier. To examine the significance of negative samples in the UDM model, we considered two different kinds of negative samples while learning.  In the first approach, we considered all samples except the positive samples as negative~\cite{silberer2016visually}. In the second, we used semantic similarity measures over the descriptions~\cite{PillaiRSSws2017Negatives}. For this, we treat a concatenation of all the descriptions associated with one object as documents, and convert these `descriptive documents' into vector space using the Distributed Memory Model of Paragraph Vectors (PV-DM)~\cite{mikolov2013distributed}. As semantically similar documents will have similar representations in vector space, we use cosine similarity to find the most distant paragraph vectors, and select the respective object instances as negative examples for our concept.


\pseudosection{UDM Structure}
We experimented with latent embedding lengths (size of $Z$) ranging from 12 to 100; in early development, we found 50 to yield the best results. This computed $Z$ formed the input features for the discriminative classifier. For the variational autoencoder, we experimented with a single hidden layer MLP, with hidden dimension ranging from 100 to 600; 500 yielded the best results.
\subsection{Experimental Setup}
\label{sec:expt:setup}
\pseudosection{Baselines}\label{sec:expt:setup:baselines} RGB-D visual classifiers are compared with two baselines. First, in the `predefined' category classifier~\cite{PillaiAAAI2018}, visual classifiers are trained for every concept and feature category: for example, ``arch'' has associated ``arch-as-color,'' ``arch-as-shape,'' and ``arch-as-object'' classifiers. The second baseline is a `category-free' approach, where logistic regression classifiers are trained for every concept with the concatenated feature set. Category-free logistic regression uses the concatenated $X$ features instead of the $Z$ features that UDM uses. Here, ``arch'' is trained as ``arch-classifier,'' accepting a concatenated set of all features as its input.

\pseudosection{Metrics and Rigor} In keeping with prior work, we measure success in grounded concept prediction via the classification performance of the learned concept classifiers. We also use the same label selection process as \cite{PillaiAAAI2018}: for each learned classifier, we selected 3--4 positive and 4--6 negative images from the test set. If the predicted probability for a test image is above 0.5 it is considered a positive result. Due to the comparatively small sizes of the datasets, we use four-fold cross-validation, and within each fold, we calculated the average F1-score across 10 trials. we ran the experiments on K20 GPUs, and jobs required no more than 6 GB of memory.

\subsection{Limited Resource Classifications}\label{subsec:overall_comparison}

\begin{table}[hbt]
\begin{tabular}{l|c|c|c|c|c|}
\cline{2-6}
 & \multirow{2}{*}{\textbf{\begin{tabular}[c]{@{}c@{}}Predefined\\ category\\ classifer\end{tabular}}} & \multirow{2}{*}{\textbf{\begin{tabular}[c]{@{}c@{}}Category-\\ free logistic\\ regression\end{tabular}}} & \multicolumn{3}{c|}{\textbf{\begin{tabular}[c]{@{}c@{}}Unified \\ discriminative method\end{tabular}}} \\ \cline{4-6} 
 &  &  & \textbf{Dim} & \textbf{Dim} & \textbf{Dim} \\
 & \multicolumn{1}{l|}{} & \multicolumn{1}{l|}{} & \textbf{12} & \textbf{50} & \textbf{100} \\ \hline
\multicolumn{1}{|l|}{\textbf{Minimum}} & 0.246 & 0.233 & 0.257 & \textbf{0.456} & 0.242 \\ \hline
\multicolumn{1}{|l|}{\textbf{Mean}} & 0.706 & 0.607 & 0.659 & \textbf{0.713} & 0.634 \\ \hline
\multicolumn{1}{|l|}{\textbf{Maximum}} & 0.956 & 0.888 & \textbf{0.968} & 0.963 & 0.900 \\ \hline
\end{tabular}

\caption{Overall summary of the F1-score distribution comparisons of all concepts. The minimum, mean and the maximum of our method are higher than all baselines, with the UDM with 50 latent dimensions showing better learning especially for difficult categories.}
\label{tab:boxplot_summary}
\end{table}

\Cref{tab:boxplot_summary} shows the overall summary of the distributional comparison of baselines with discriminative model variants. Here we used RGB-D visual features from 72 objects for the analysis. Our method outperforms our baselines. This shows the classification performance improvement of UDM method even for highly noisy, visually varied, and underperformed visual classifiers compared to the well known predefined visual classifier baseline.  

\begin{figure}[bt]
\centering
\includegraphics[trim={0cm 9 0 0},clip]{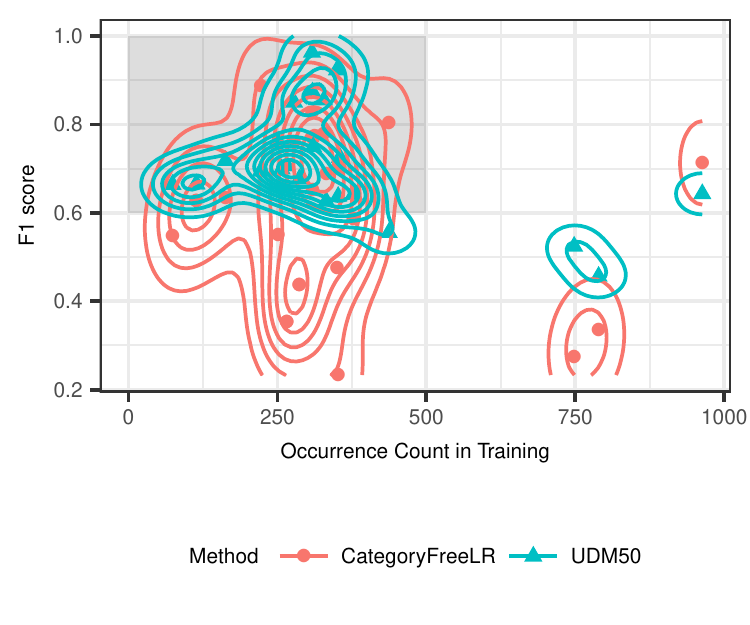}
\caption{ %
The comparison of the F1-score distribution of all concepts of the unified discriminative method vs. category free logistic regression. The goal is a high F1 score with a smaller number of occurrences, so the upper left quadrant (shaded) is the target. F1-score performance of UDM is both high and consistent with limited annotation. %
}
\label{fig:f1score_comparison_plot}
\end{figure}

In addition to this strong performance, UDM improves on the logistic regression-based category-agnostic baseline for concepts that have fewer discriminative training instances. The vocabulary used in our dataset for certain concepts is highly varied, meaning relatively few annotations that use each linguistic token. We can see this in \cref{fig:f1score_comparison_plot}, which plots the density estimate of F1 performance for each concept classifier vs. the number of labeled examples provided for our UDM method and category-free logistic regression baseline. As our objective is to achieve good language acquisition with limited annotation, the goal is a high F1-score for learning from a small number of occurrences (upper left quadrant, shaded). The tighter density of UDM (blue triangles) shows that performance is high given limited examples.

\pseudosection{Efficacy over CNN models} Our analysis with CNN variants shows the quality improvement of UDM with CNN features, compared to CNN baselines. Here we consider the features extracted using SmallVGGNet and NasnetLarge CNN variants to test the efficacy of UDM over CNN features. The results show that the minimum F1 is improved for UDM (0.37 for SmallVGGNet, 0.30 for NasnetLarge) compared to CNN baselines (0.0 for both SmallVGGNet and Nasnetlarge). From the results, we understand that UDM is able to elevate the base quality in classification over CNN features. We notice SmallerVGGNet performs extremely well on color and object classification  (0.75--1.0), whereas it provides low scores for shape classification (0.0--0.47). Shape descriptions are more variable, and so more sparse, compared to color and object description. That results in comparatively poor classification performance from the high-dimensional CNN feature set. We also notice that NASNetLarge classifiers give very low scores in color classification, especially for concepts like \Concept{red} and \Concept{yellow}, where we have a lot of variations in the object set, while our UDM is able to extract a meaningful representation out of NASNetLarge features. 


\begin{figure}[t]
\centering
\includegraphics[width=\columnwidth]{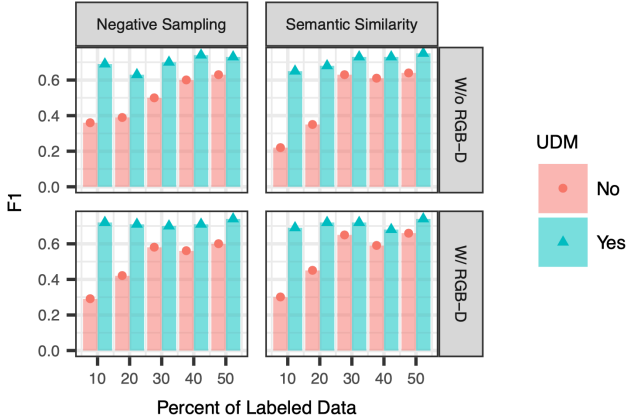}
\caption{Classification performance of UDM in different architecture variations with small training data. We consider two negative sample varieties (semantically dissimilar patterns as negative examples, and non-positive samples as negative examples), and two feature input combinations (SmallVGGNet CNN features with and without RGB-D).}
\label{figure:comparison_limiteddata}
\end{figure}

\pseudosection{Low Visual and Linguistic Resources} The discriminative model (UDM) matches descriptions to objects better than our baselines with less visual and linguistic training data. \Cref{figure:comparison_limiteddata} shows the performance comparison of UDM with baselines on limited training data with different learning parameters. With 10\% of training data, all UDM variants reach an average F1-score $\geq 0.65$, while baselines are unable to generalize the learning with limited training data. Baselines (with RGB-D) required a minimum of 30\% of training data to learn groundings for the most important object concepts, such as `banana,' `tomato,' and `lime' in the dataset. A CNN baseline (SmallVGGNet features without RGB-D) needed 40\% of the training data to learn shape concepts, such as `triangular,' `rectangular,' and `cylinder.' Sparsity in the use of particular descriptors results in an F1-score $\leq 0.5$ for baseline classification, while low-dimensional representational embeddings yield improved classification for UDM.

\begin{figure}[t]
\centering
\includegraphics[width=\columnwidth]{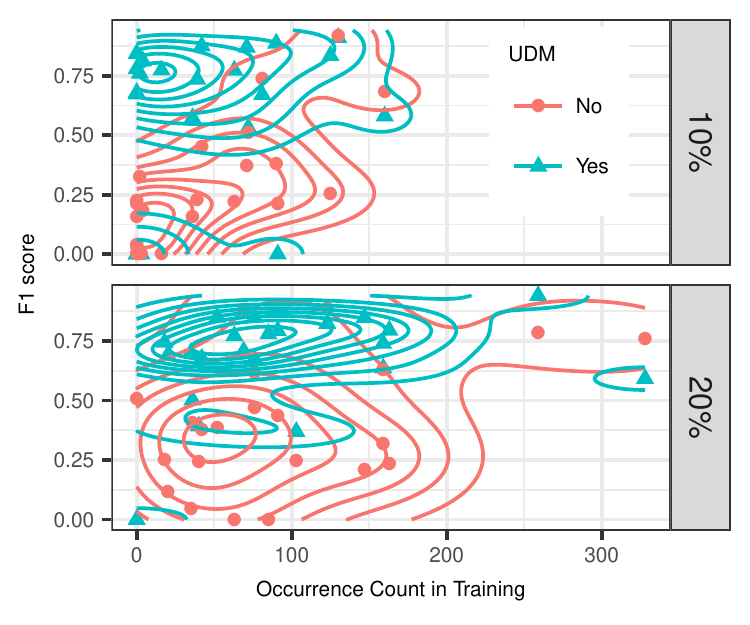}
\caption{F1-score distribution comparison of CNN variant (SmallVGGNet) vs. UDM, for all concepts with varying annotation frequency (horizontal axis). We consider setting using either 10\% or 20\% of the labeled data. The performance of UDM is high and consistent with very limited annotation.}
\label{fig:f1score_ablation_comparison_plot}
\end{figure} 

\Cref{fig:f1score_ablation_comparison_plot} shows the high quality classification performance of UDM as only a small portion of the dataset is made available for discriminative training. We extract visual features using SmallVGGNet, where \cref{fig:f1score_ablation_comparison_plot} shows the comparison of using those features with UDM vs. without (i.e., just logistic regression). Further, we show how UDM can make use of a small percentage of the total training data. These cases serve to understand the learning growth of these architectures as comparing to our UDM approaches.

\begin{figure}[t]
\centering
\includegraphics[width=\columnwidth]{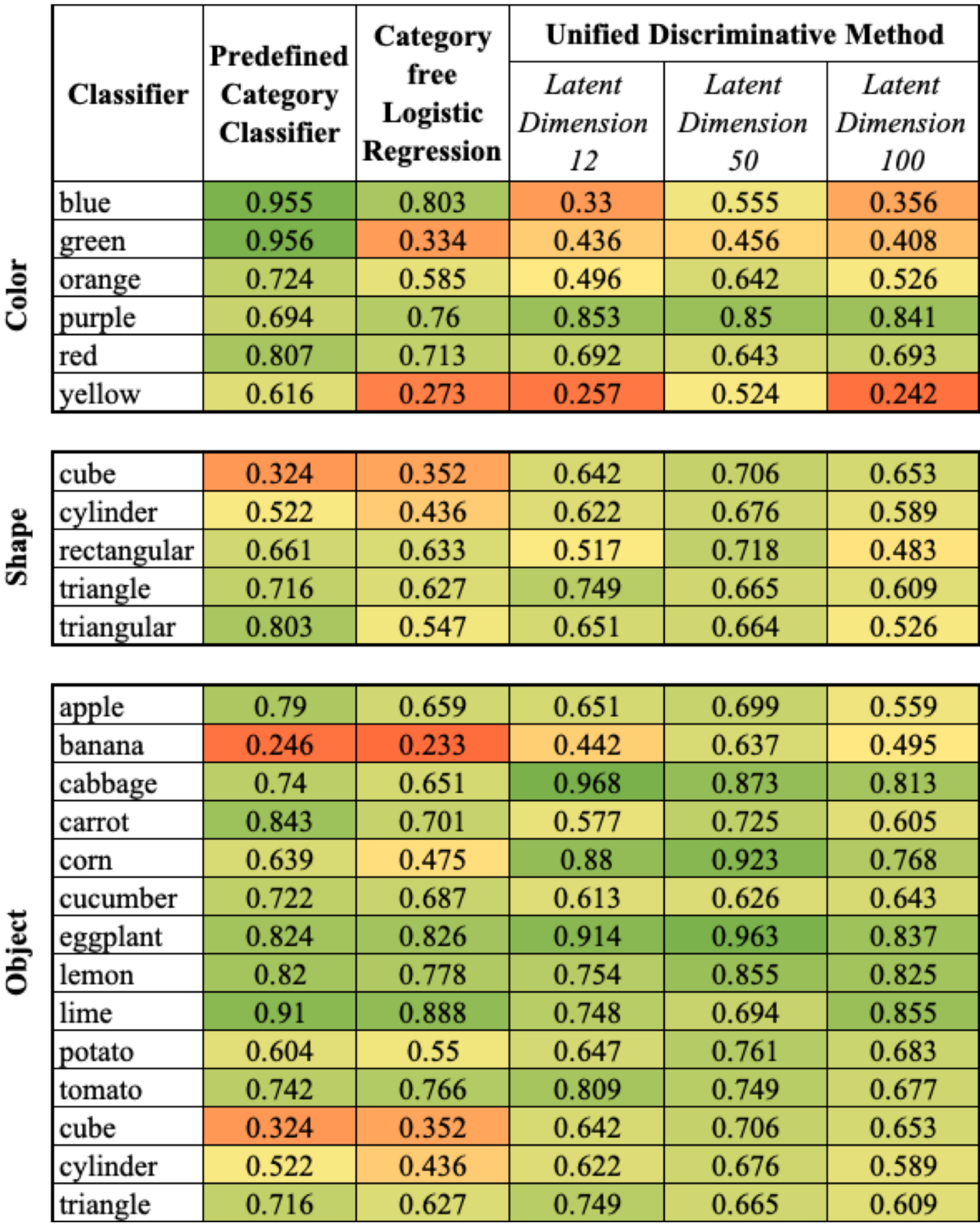}
	\caption{Averaged macro F1-score comparison of our unified discriminative method against other approaches for every concept  with RGB-D features. We segment the classifiers by category here for ease of analysis: our UDM models do not consider category types. UDM with latent dimension 50 is able to provide promising performance in grounded language acquisition for all categories. Color-specific visual classifiers perform better compared to the category free logistic regression baseline. Object and shape classifiers perform well with our method (UDM) with latent dimension 50 compared to other approaches.}
    \label{fig:overall_f1score}
\end{figure}

\pseudosection{Concept-wise Comparison} 
Figure \ref{fig:overall_f1score} shows the performance comparison of two baselines (see section \ref{sec:expt:setup:baselines}) and the variants of the unified discriminative method for every meaningful concept for the RGB-D dataset. The predefined category-specific baseline grounds color specific language ``terms'' exceptionally well compared to other approaches. On average, color classifiers had an F1-score of 0.792 for the predefined category classifier, 0.578 for category-free logistic regression, and 0.611 for UDM with latent dimension 50. However, our method with latent dimension 50 is able to perform better than the category-free logistic regression where classifier input is accepted as a vector of raw features. 
Our method with latent dimension 50 outperforms both baselines for shape classification, with an average F1-score of 0.69, where category-free logistic regression scored 0.52, and category-specific approach scored 0.61.  In the case of object classification, which is comparatively complex, our method with latent dimensions 50 performs better compared to predefined category classifier and category-free logistic regression. F1 scores for all methods are as follows: Predefined category classifier, 0.674; category-free logistic regression, 0.616; and UDM with latent dimension 50, 0.754. When the minimum F1-score for UDM with dimension 50 is 0.626, the baseline predefined category classifier's peak F1 is as low as 0.246 and category free logistic regression F1 equals 0.233.  This verifies the quality enhancement in visual classification for limited data settings. Overall, UDM achieves 0.7218 micro averaged F1 score and the category-free approach achieves 0.6699; meanwhile, UDM outperforms the previously-published, predefined category classifier (0.7192). This shows that UDM performs as strong as the predefined classifier, and eliminates the need for creating separate category-specific classifiers. 

\begin{figure}[t]
\centering
\includegraphics[width=\columnwidth]{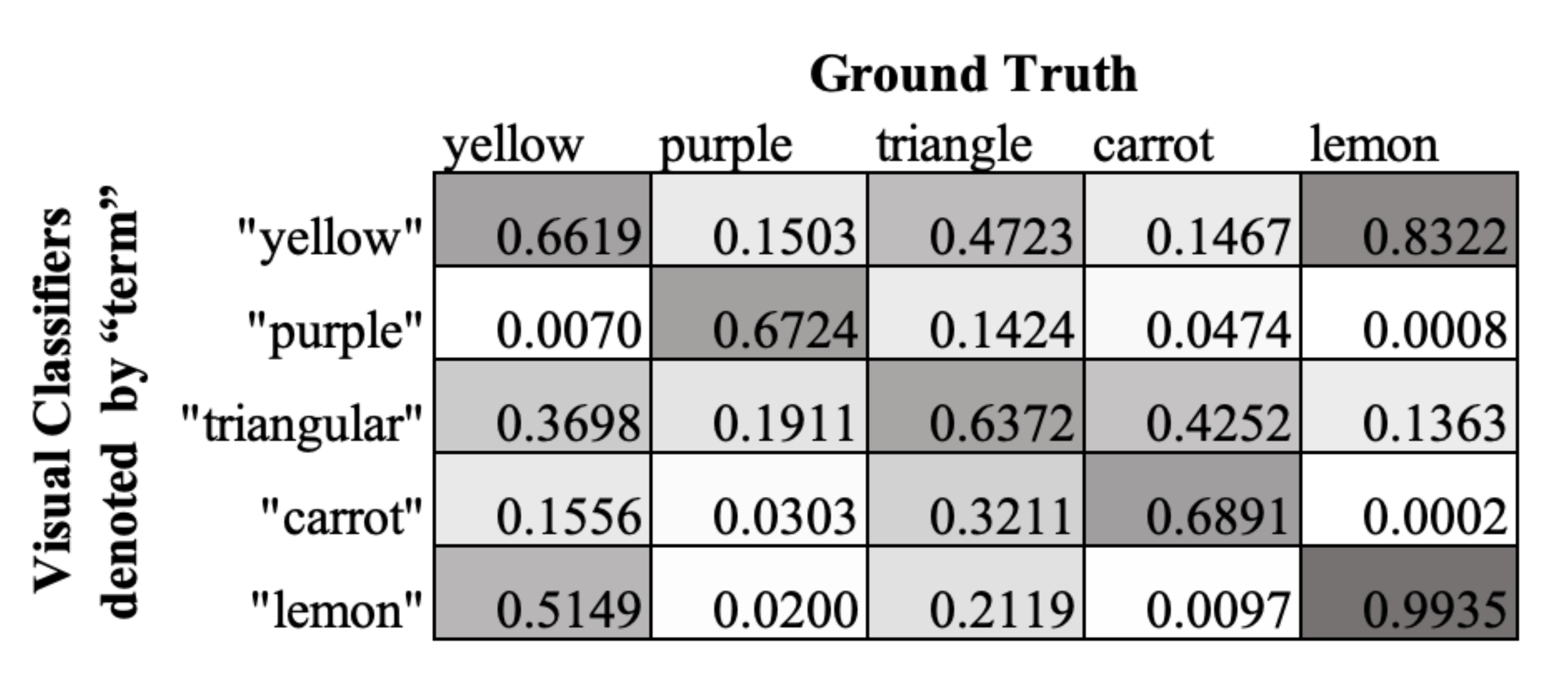}
	\caption{Prediction probabilities of selected visual classifiers (\textit{y}-axis) against ground truth objects (\textit{x}-axis) selected from a held-out test set with RGB-D features. This confusion matrix exhibits the prediction confidence of the unified discriminative method (UDM) run against real-world objects. Color, shape, and object variations add complexity to the performance.}
    \label{fig:prob_matrix}
\end{figure}

\pseudosection{Language Prediction Probabilities} \Cref{fig:prob_matrix} shows the association between visual classifiers and the ground truth after learning the language and vision components through our unified discriminative method. Color classifiers show strong performance.
In this dataset, ``yellow'' objects ranged from bananas to corn, while ``purple'' objects were limited to eggplant, plum, and cabbage. 

Compared to color classifiers, object classifiers are able to predict object instances with great prediction strength. The ``lemon'' classifier shows the positive association with `yellow objects, and strong predictive ability on a lemon. The shape features of a carrot are complex compared to a lemon, so it is unsurprising that the predictive power of the learned ``carrot'' classifier is not strong compared to a ``lemon'' classifier. From different angles, pictures of carrots show very different shapes, while lemons are almost the same from all angles (e.g., the angle of a carrot's position made it look like a triangle). From an elevated view, the angle of the carrot's position made it look like the side of a triangle in our pictures. The complexity of the features affects the classification accuracy substantially.

\pseudosection{Concept-wise Classification Analysis} \label{subsec:analysis} %
From \cref{fig:overall_f1score}, we see in general that color concepts are learned well by the predefined category classifier. The predefined category classifier approach uses \textit{selective}, low-dimensional, low variability training features, defined specifically for each category, which leads to better classification. Low-dimensional features lack representation compared to complex high-dimensional features when all features are combined into representational embeddings. In our experiments, UDM combines lower-variation color features with other, higher variability shape/object features. This combination leads to lower UDM performance on color concepts. For example, with the ``blue'' concept, less variability in the training features and less noise in the annotation (such as what items annotators described as ``blue'') yielded a higher-quality predefined category classifier, but the addition of highly varied shape features confused the UDM classifier. The case is similar in case of green and orange concepts. In the case of ``red,'' high variability in the dataset and moderate noise in the annotation played a significant role in classification. Similarly, annotators describe a variety of vegetables such as corn, potato, orange, and banana as ``yellow'' concepts. This gives a highly variable visual training data to the classifier, and UDM performance is affected. Overall, our qualitative analysis suggests that less annotation variability seems to be a key component in the UDM concept grounding.

Shape classification presents a different picture. Even with highly variable shape annotations, UDM extracts useful information from the training data and performs better than the traditional grounding methods. Since the shape information dominates in the cumulative training representation, UDM obtains useful information and results in stronger classification compared to the traditional grounding methods. Even when the annotation is not noisy, but the visual complexity is high, UDM presents a reliable classification. In the dataset, ``cylinder'' and ``semi-cylinder'' objects are annotated as cylinders, and it makes the annotation for the ``cylinder'' concept less noisy. In the case of ``rectangular'' concepts, annotation is highly noisy. Many children's toys are described as rectangular, including the arch, triangle, and cylinder objects. High noise increases the complexity in the training data, yet UDM can partially account for this complexity. The objects with complicated shapes such as triangles are difficult in visual classification. Extracting essential bits from the complex training structure is a challenge.

Noise in annotations affects the quality of the performance of ``object'' concepts as well. Plums, potatoes, tomatoes, oranges, and limes are occasionally described as apples, leading to highly varied visuals in the training data. Meanwhile, the consistently high-quality annotations of ``banana'' yield a robust classifier. 
It is interesting to note that bananas are sometimes confused with cucumbers, especially since some of the bananas in our data are green. Though the shape features are not identical, the cucumber is visually similar. 


\subsection{Multi-Lingual Verifications}\label{subsec:multilingual}
Our objective here is to show this simple discriminative method is generalizable to multi-lingual visual classification (see~\cref{tab:multi-lang_summary}). We use Spanish and Hindi descriptions~\cite{Kery2019ROMAN} collected from non-trained humans for 72 RGB-D object  dataset~\cite{PillaiAAAI2018} for this experiment. The language dataset contains 5,100 Spanish and 5,700 Hindi descriptions. While the Spanish language dataset contains 35 color, 51 shape, and 138 object concepts, the Hindi descriptions include 25 color, 34 shape, and 135 object concepts. Shape and object concepts in both the datasets are highly varied and diverse, causing the classification to be difficult. Color concepts in the Spanish set seem concise and less varied, but gender-based inflectional differences in the description cause the color concepts diverse in the Hindi set. Because the current top performance on this Hindi and Spanish grounding problem uses logistic regression, we use category free logistic regression as a baseline here. With both Spanish and Hindi descriptions, UDM achieves consistent performance improvements compared to the baseline.  This verification validates that our category-free learning approach is useful across languages and complexity. 
\begin{table}
\small
\centering
\resizebox{\columnwidth}{!}{
\renewcommand{\arraystretch}{1.4}
\begin{tabular}{|c|c|c|c|c|c|c|c|c|}
\hline
\textbf{Language} & \textbf{Sampling} & \textbf{10\%} & \textbf{20\%} & \textbf{30\%}  & \textbf{40\%}  & \textbf{50\%}  & \textbf{60\%}  & \textbf{70\%}  \\ \hline
\multirow{2}{*}{\textbf{Spanish}} & Category-free LR &  0.05 &	0.14 &	0.23 &	0.41 &	0.43 &	0.49 & 0.48\\ \cline{2-9}
 & \bf{UDM} & \bf{0.14} &	\bf{0.24}&	\bf{0.32}&	\bf{0.45}&	\bf{0.45}&	\bf{0.48}&	\bf{0.52}\\ \hline
 
\multirow{2}{*}{\textbf{Hindi}} & Category-free LR &  0.038&	0.160&	0.228&	0.334&	0.437&	0.504&	0.518\\ \cline{2-9}
 & \bf{UDM} & \bf{0.187}&	\bf{0.290}&	\bf{0.413}&	\bf{0.490}&	\bf{0.516}&	\bf{0.536}&	\bf{0.552} \\ \hline 
\end{tabular} %
}
\caption{F1-score performance of UDM in multi-lingual classification with less training data. UDM gives a consistent improvement compared to the category free logistic regression baseline with both Spanish and Hindi training data. }
\label{tab:multi-lang_summary}
\end{table}

\subsection{Highly Complex, Multi-Colored Resource Verification}\label{subsec:high_resource}
In this experiment, we used an RGB-D dataset with 300 objects for testing. As a baseline, we used the 1024 dimension feature set extracted using NASNetLarge, a CNN variant, for grounding natural language concepts. Even with 10\% of total training data, we notice our discriminative method performs better than the CNN approach. When our approach scored 0.46 F1-score across all concepts, NASNetLarge only scored 0.39. This shows our approach is effective in learning a better representative embedding from the visual features and is generalizable to other datasets. 




\section{Conclusion}\label{subsec:conclusion}


We have presented a simple, strong approach for learning a unified language grounding model that is not constrained to predefined category attributes. We show that pre-training a straightforward Gaussian variational autoencoder efficiently grounds linguistic concepts found in unconstrained natural language to real sensor data. To compare against previous, more limited work, our evaluation primarily focuses on prediction of color, shape, and object descriptions. We also present experimental results demonstrating successful learning of a broad range of concepts from a well-studied RGB+D dataset. We hope that our improvements in low-resource settings will provide tools and insights for future work.





{
\fontsize{9.0pt}{10.0pt} \selectfont 
\bibliographystyle{IEEEtran}
\bibliography{main}

\begin{thebibliography}{10}
\providecommand{\url}[1]{#1}
\csname url@rmstyle\endcsname
\providecommand{\newblock}{\relax}
\providecommand{\bibinfo}[2]{#2}
\providecommand\BIBentrySTDinterwordspacing{\spaceskip=0pt\relax}
\providecommand\BIBentryALTinterwordstretchfactor{4}
\providecommand\BIBentryALTinterwordspacing{\spaceskip=\fontdimen2\font plus
\BIBentryALTinterwordstretchfactor\fontdimen3\font minus
  \fontdimen4\font\relax}
\providecommand\BIBforeignlanguage[2]{{%
\expandafter\ifx\csname l@#1\endcsname\relax
\typeout{** WARNING: IEEEtran.bst: No hyphenation pattern has been}%
\typeout{** loaded for the language `#1'. Using the pattern for}%
\typeout{** the default language instead.}%
\else
\language=\csname l@#1\endcsname
\fi
#2}}

\bibitem{harnad1990symbol}
S.~Harnad, ``The symbol grounding problem,'' \emph{Physica D: Nonlinear
  Phenomena}, 1990.

\bibitem{chi2020aaai}
T.-C. Chi, M.~Shen, M.~Eric, S.~Kim, and D.~Hakkani-Tur, ``Just ask: An
  interactive learning framework for vision and language navigation,'' in
  \emph{Conference on Artificial Intelligence (AAAI)}, 2020.

\bibitem{wang2019hierarchical}
B.~Wang, L.~Ma, W.~Zhang, W.~Jiang, and F.~Zhang, ``Hierarchical photo-scene
  encoder for album storytelling,'' in \emph{Conference on Artificial
  Intelligence (AAAI)}, 2019.

\bibitem{sinapov2016IJCAI}
J.~Sinapov, P.~Khante, M.~Svetlik, and P.~Stone, ``Learning to order objects
  using haptic and proprioceptive exploratory behaviors,'' in \emph{Int'l Joint
  Conf. on Artificial Intelligence (IJCAI)}, 2016.

\bibitem{Antol_2015_ICCV}
S.~Antol, A.~Agrawal, J.~Lu, M.~Mitchell, D.~Batra, C.~Lawrence~Zitnick, and
  D.~Parikh, ``Vqa: Visual question answering,'' in \emph{ICCV}, 2015.

\bibitem{MatuszekICML2012}
C.~Matuszek, N.~I. J.~C. on~Artificial Intelligence~(IJCAI)Gerald,
  L.~Zettlemoyer, L.~Bo, and D.~Fox, ``{A Joint Model of Language and
  Perception for Grounded Attribute Learning},'' in \emph{Int'l Conf. on
  Mahcine Learning}, 2012.

\bibitem{richards2020IROS}
L.~E. Richards, K.~Darvish, and C.~Matuszek, ``Learning object attributes with
  category-free grounded language from deep featurization,'' in
  \emph{Intelligent Robots and Systems}, 2020.

\bibitem{bisk2020experience}
Y.~Bisk, A.~Holtzman, J.~Thomason, J.~Andreas, Y.~Bengio, J.~Chai, M.~Lapata,
  A.~Lazaridou, J.~May, A.~Nisnevich, \emph{et~al.}, ``Experience grounds
  language,'' in \emph{EMNLP}, 2020.

\bibitem{doi:10.1080/01691864.2016.1164622}
T.~Taniguchi, T.~Nagai, T.~Nakamura, N.~Iwahashi, T.~Ogata, and H.~Asoh,
  ``Symbol emergence in robotics: a survey,'' \emph{Advanced Robotics}, 2016.

\bibitem{BergBergShih2010}
T.~Berg, A.~Berg, and J.~Shih, ``Automatic attribute discovery and
  characterization from noisy web data,'' \emph{European Conference on Computer
  Vision}, 2010.

\bibitem{Kery2019ROMAN}
C.~Kery, N.~Pillai, C.~Matuszek, and F.~Ferraro, ``Building language-agnostic
  grounded language learning systems,'' in \emph{Ro-Man}, 2019.

\bibitem{PillaiRSS2016workshopActive}
N.~Pillai, K.~K. Budhraja, and C.~Matuszek, ``Improving grounded language
  acquisition efficiency using interactive labeling,'' in \emph{R:SS Workshop
  on MLHRC}, 2016.

\bibitem{paul2016efficient}
R.~Paul, J.~Arkin, N.~Roy, and T.~M~Howard, ``Efficient grounding of abstract
  spatial concepts for natural language interaction with robot manipulators,''
  in \emph{Robotic Science and Systems}, 2016.

\bibitem{tellex2011approaching}
S.~Tellex, T.~Kollar, S.~Dickerson, M.~R. Walter, A.~G. Banerjee, S.~Teller,
  and N.~Roy, ``Approaching the symbol grounding problem with probabilistic
  graphical models,'' \emph{AI Magazine}, 2011.

\bibitem{brawer2018situated}
J.~Brawer, O.~Mangin, A.~Roncone, S.~Widder, and B.~Scassellati, ``Situated
  human--robot collaboration: predicting intent from grounded natural
  language,'' in \emph{Intelligent Robots and Systems}, 2018.

\bibitem{Akula2020WordsAE}
A.~R. Akula, S.~Gella, Y.~Al-Onaizan, S.-C. Zhu, and S.~Reddy, ``Words aren't
  enough, their order matters: On the robustness of grounding visual referring
  expressions,'' in \emph{Association for Computational Linguistics}, 2020.

\bibitem{rao2018learning}
A.~B. Rao, K.~Krishnan, and H.~He, ``Learning robotic grasping strategy based
  on natural-language object descriptions,'' in \emph{Intelligent Robots and
  Systems}, 2018.

\bibitem{levine2018learning}
S.~Levine, P.~Pastor, A.~Krizhevsky, J.~Ibarz, and D.~Quillen, ``Learning
  hand-eye coordination for robotic grasping with deep learning and large-scale
  data collection,'' \emph{IJRR}, 2018.

\bibitem{thomason2016IJCAI}
J.~Thomason, J.~Sinapov, M.~Svetlik, P.~Stone, and R.~J. Mooney, ``Learning
  multi-modal grounded linguistic semantics by playing {``I Spy''},'' in
  \emph{Int'l Joint Conf. on Artificial Intelligence (IJCAI)}, 2016.

\bibitem{Nguyen2020RobotOR}
T.~Nguyen, N.~Gopalan, R.~Patel, M.~Corsaro, E.~Pavlick, and S.~Tellex, ``Robot
  object retrieval with contextual natural language queries,'' \emph{Robotics
  Science and Systems}, 2020.

\bibitem{Chen_2020_CVPR}
R.~Chen, W.~Huang, B.~Huang, F.~Sun, and B.~Fang, ``Reusing discriminators for
  encoding: Towards unsupervised image-to-image translation,'' in
  \emph{Computer Vision and Pattern Recognition}, 2020.

\bibitem{pitti2021brain}
A.~Pitti, M.~Quoy, S.~Boucenna, and C.~Lavandier, ``Brain-inspired model for
  early vocal learning and correspondence matching using free-energy
  optimization,'' \emph{PLoS Computational Biology}, 2021.

\bibitem{Liu2021TaskAG}
Z.~Liu, Y.~Li, L.~Yao, X.~Wang, and G.~Long, ``Task aligned generative
  meta-learning for zero-shot learning,'' in \emph{Conference on Artificial
  Intelligence (AAAI)}, 2021.

\bibitem{boudiaf2020information}
M.~Boudiaf, I.~Ziko, J.~Rony, J.~Dolz, P.~Piantanida, and I.~Ben~Ayed,
  ``Information maximization for few-shot learning,'' \emph{Neural Information
  Processing Systems}, 2020.

\bibitem{DBLP:conf/naacl/DevlinCLT19}
J.~Devlin, M.-W. Chang, K.~Lee, and K.~Toutanova, ``Bert: Pre-training of deep
  bidirectional transformers for language understanding,'' in \emph{NAACL-HLT},
  2019.

\bibitem{lu2019vilbert}
J.~Lu, D.~Batra, D.~Parikh, and S.~Lee, ``Vilbert: Pretraining task-agnostic
  visiolinguistic representations for vision-and-language tasks,'' in
  \emph{Neural Information Processing Systems}, 2019.

\bibitem{lei2020mart}
J.~Lei, L.~Wang, Y.~Shen, D.~Yu, T.~L. Berg, and M.~Bansal, ``Mart:
  Memory-augmented recurrent transformer for coherent video paragraph
  captioning,'' in \emph{Association for Computational Linguistics}, 2020.

\bibitem{Tan_2018_CVPR}
Q.~Tan, L.~Gao, Y.-K. Lai, and S.~Xia, ``Variational autoencoders for deforming
  3d mesh models,'' in \emph{Computer Vision and Pattern Recognition}, June
  2018.

\bibitem{yamada2018paired}
T.~Yamada, H.~Matsunaga, and T.~Ogata, ``Paired recurrent autoencoders for
  bidirectional translation between robot actions and linguistic
  descriptions,'' \emph{IEEE Robotics and Automation Letters}, vol.~3, no.~4,
  pp. 3441--3448, 2018.

\bibitem{Qiao_2020_CVPR}
Z.~Qiao, Y.~Zhou, D.~Yang, Y.~Zhou, and W.~Wang, ``Seed: Semantics enhanced
  encoder-decoder framework for scene text recognition,'' in \emph{Computer
  Vision and Pattern Recognition}, 2020.

\bibitem{silberer2014learning}
C.~Silberer and M.~Lapata, ``Learning grounded meaning representations with
  autoencoders,'' in \emph{Association for Computational Linguistics}, 2014.

\bibitem{rohrbach2016grounding}
A.~Rohrbach, M.~Rohrbach, R.~Hu, T.~Darrell, and B.~Schiele, ``Grounding of
  textual phrases in images by reconstruction,'' in \emph{European Conference
  on Computer Vision}, 2016.

\bibitem{PillaiAAAI2018}
N.~Pillai and C.~Matuszek, ``Unsupervised end-to-end data selection for
  grounded language learning,'' in \emph{Conference on Artificial Intelligence
  (AAAI)}, 2018.

\bibitem{nyga2017no}
D.~Nyga, M.~Picklum, and M.~Beetz, ``What no robot has seen
  before—probabilistic interpretation of natural-language object
  descriptions,'' in \emph{{International Conference on Robotics and
  Automation}}, 2017.

\bibitem{Hariharan_2017_ICCV}
B.~Hariharan and R.~Girshick, ``Low-shot visual recognition by shrinking and
  hallucinating features,'' in \emph{ICCV}, Oct 2017.

\bibitem{vinyals2016matching}
O.~Vinyals, C.~Blundell, T.~Lillicrap, D.~Wierstra, \emph{et~al.}, ``Matching
  networks for one shot learning,'' in \emph{Neural Information Processing
  Systems}, 2016.

\bibitem{liu-etal-2019-second}
Q.~Liu, D.~McCarthy, and A.~Korhonen, ``Second-order contexts from lexical
  substitutes for few-shot learning of word representations,'' in
  \emph{*{SEM}}, 2019.

\bibitem{sun-etal-2019-hierarchical}
S.~Sun, Q.~Sun, K.~Zhou, and T.~Lv, ``Hierarchical attention prototypical
  networks for few-shot text classification,'' in \emph{EMNLP}, 2019.

\bibitem{lampert2014attribute}
C.~H. Lampert, H.~Nickisch, and S.~Harmeling, ``Attribute-based classification
  for zero-shot visual object categorization,'' in \emph{PAMI}, 2014.

\bibitem{Han_2020_CVPR}
Z.~Han, Z.~Fu, and J.~Yang, ``Learning the redundancy-free features for
  generalized zero-shot object recognition,'' in \emph{Computer Vision and
  Pattern Recognition}, 2020.

\bibitem{Wang_2020_CVPR}
S.~Wang, K.-H. Yap, J.~Yuan, and Y.-P. Tan, ``Discovering human interactions
  with novel objects via zero-shot learning,'' in \emph{Computer Vision and
  Pattern Recognition}, 2020.

\bibitem{kingma2014semi}
D.~P. Kingma, S.~Mohamed, D.~J. Rezende, and M.~Welling, ``Semi-supervised
  learning with deep generative models,'' in \emph{Neural Information
  Processing Systems}, 2014.

\bibitem{lai2011dataset}
K.~Lai, L.~Bo, X.~Ren, and D.~Fox, ``A large-scale hierarchical multi-view
  {RGB-D} object dataset,'' in \emph{ICRA}, 2011.

\bibitem{BoLaiRenEtAl2011}
L.~Bo, K.~Lai, X.~Ren, and D.~Fox, ``Object recognition with hierarchical
  kernel descriptors,'' in \emph{Computer Vision and Pattern Recognition},
  2011.

\bibitem{deng09}
J.~Deng, W.~Dong, R.~Socher, L.-J. Li, K.~Li, and L.~Fei-Fei, ``Imagenet: A
  large-scale hierarchical image database,'' in \emph{CVPR}, 2009.

\bibitem{DBLP:journals/corr/SimonyanZ14a}
K.~Simonyan and A.~Zisserman, ``Very deep convolutional networks for
  large-scale image recognition,'' in \emph{ICLR}, 2015.

\bibitem{zoph2018learning}
B.~Zoph, V.~Vasudevan, J.~Shlens, and Q.~V. Le, ``Learning transferable
  architectures for scalable image recognition,'' in \emph{Computer Vision and
  Pattern Recognition}, 2018.

\bibitem{silberer2016visually}
C.~Silberer, V.~Ferrari, and M.~Lapata, ``Visually grounded meaning
  representations,'' \emph{PAMI}, 2016.

\bibitem{PillaiRSSws2017Negatives}
N.~Pillai and C.~Matuszek, ``Identifying negative exemplars in grounded
  language data sets,'' in \emph{R:SS Workshop on Spatial-Semantic
  Representations in Robotics}, 2017.

\bibitem{mikolov2013distributed}
T.~Mikolov, I.~Sutskever, K.~Chen, G.~S. Corrado, and J.~Dean, ``Distributed
  representations of words and phrases and their compositionality,'' in
  \emph{Neural Information Processing Systems}, 2013.

\bibitem{Mooney2008}
R.~J. Mooney, ``Learning to connect language and perception,'' in
  \emph{Conference on Artificial Intelligence (AAAI)}, 2008.

\bibitem{hatori2018interactively}
J.~Hatori, Y.~Kikuchi, S.~Kobayashi, K.~Takahashi, Y.~Tsuboi, Y.~Unno, W.~Ko,
  and J.~Tan, ``Interactively picking real-world objects with unconstrained
  spoken language instructions,'' in \emph{{International Conference on
  Robotics and Automation}}, 2018.

\bibitem{ren2015faster}
S.~Ren, K.~He, R.~Girshick, and J.~Sun, ``Faster r-cnn: Towards real-time
  object detection with region proposal networks,'' in \emph{Neural Information
  Processing Systems}, 2015.

\bibitem{karpathy2014large}
A.~Karpathy, G.~Toderici, S.~Shetty, T.~Leung, R.~Sukthankar, and L.~Fei-Fei,
  ``Large-scale video classification with convolutional neural networks,'' in
  \emph{Computer Vision and Pattern Recognition}, 2014.

\bibitem{Isola_2017_CVPR}
P.~Isola, J.-Y. Zhu, T.~Zhou, and A.~A. Efros, ``Image-to-image translation
  with conditional adversarial networks,'' in \emph{Computer Vision and Pattern
  Recognition}, 2017.

\bibitem{plappert2018learning}
M.~Plappert, C.~Mandery, and T.~Asfour, ``Learning a bidirectional mapping
  between human whole-body motion and natural language using deep recurrent
  neural networks,'' \emph{RAS}, 2018.

\bibitem{Vasudevan_2018_CVPR}
A.~Balajee~Vasudevan, D.~Dai, and L.~Van~Gool, ``Object referring in videos
  with language and human gaze,'' in \emph{Computer Vision and Pattern
  Recognition}, 2018.

\bibitem{Wang_2018_CVPR}
Y.~Wang, J.~van~de Weijer, and L.~Herranz, ``Mix and match networks:
  Encoder-decoder alignment for zero-pair image translation,'' in
  \emph{Computer Vision and Pattern Recognition}, June 2018.

\bibitem{wan2017crossing}
C.~Wan, T.~Probst, L.~Van~Gool, and A.~Yao, ``Crossing nets: Combining gans and
  vaes with a shared latent space for hand pose estimation,'' in \emph{Computer
  Vision and Pattern Recognition}, 2017.

\bibitem{ahn2018text2action}
H.~Ahn, T.~Ha, Y.~Choi, H.~Yoo, and S.~Oh, ``Text2action: Generative
  adversarial synthesis from language to action,'' in \emph{{International
  Conference on Robotics and Automation}}, 2018.

\bibitem{mao2016image}
X.~Mao, C.~Shen, and Y.-B. Yang, ``Image restoration using very deep
  convolutional encoder-decoder networks with symmetric skip connections,'' in
  \emph{Neural Information Processing Systems}, 2016.

\bibitem{cadena2016multi}
C.~Cadena, A.~R. Dick, and I.~D. Reid, ``Multi-modal auto-encoders as joint
  estimators for robotics scene understanding.'' in \emph{Robotic Science and
  Systems}, 2016.

\bibitem{Elhoseiny_2013_ICCV}
M.~Elhoseiny, B.~Saleh, and A.~Elgammal, ``Write a classifier: Zero-shot
  learning using purely textual descriptions,'' in \emph{ICCV}, December 2013.

\end{thebibliography}
}
\nocite{harnad1990symbol,Mooney2008,hatori2018interactively, ren2015faster,karpathy2014large,Isola_2017_CVPR,plappert2018learning,Vasudevan_2018_CVPR,Wang_2018_CVPR,wan2017crossing,ahn2018text2action,mao2016image,cadena2016multi,Elhoseiny_2013_ICCV}

\end{document}